\documentclass[runningheads]{llncs}
 
\usepackage{eccv}

\usepackage{eccvabbrv}

\usepackage{graphicx}
\usepackage{booktabs}

\usepackage{multirow}
\usepackage{arydshln} 
\usepackage[table]{xcolor} 
\usepackage{makecell}
\usepackage{colortbl}
\usepackage{pifont}
\usepackage{amssymb}   
\usepackage{pifont}    
\usepackage{comment}
\usepackage{placeins}   

\usepackage[accsupp]{axessibility}  

\usepackage{hyperref}

\usepackage{orcidlink}
\usepackage{tikz}
\usepackage{pgfplots}
\pgfplotsset{compat=1.18}
\newcommand{\ours}{PANY}

\newcommand{\cmark}{\ding{51}}
\newcommand{\xmark}{\ding{55}}

\title{Pose Anything Anywhere: Model-free Object Poses from Arbitrary References
}
\titlerunning{Pose Anything Anywhere}

\author{
Hongli Xu$^{*1}$,
Jiaqi Hu$^{*1,3}$,
Junwen Huang$^{*\dagger1,2}$,
Boyang Zhong$^{1}$,\\
Peter KT Yu$^{4,5}$,
Nassir Navab$^{1,2}$,
Benjamin Busam$^{1,2}$,
Slobodan Ilic$^{1,3}$\\[2mm]
{\small $^*$ Equal contribution $^\dagger$ Corresponding author}
}

\authorrunning{H. Xu et al.}

\institute{
$^{1}$Technical University of Munich \quad 
$^{2}$Munich Center for Machine Learning (MCML) \quad \\
$^{3}$Siemens AG \quad  
$^{4}$XYZ Robotics \quad
$^{5}$ROBOX \quad
}

\begin{document}

\maketitle
\footnotetext{* Equal contribution. The first three authors are listed in random order.}

\begin{abstract}
Estimating the 6D pose of unseen objects is a fundamental yet challenging problem for open-world robotics and embodied perception. Model-based methods are accurate but depend on CAD assets or heavy onboarding, while most model-free approaches are still limited to pairwise single-anchor matching and thus fail under occlusion and large viewpoint changes with low query--reference overlap. Therefore, we present \textbf{\ours{}}, a unified model-free framework that seamlessly supports both RGB and RGB-D inputs, operates on one or sparse pose-free reference views, and generalizes effectively to novel objects. Built on a multi-view transformer geometry backbone, \ours{} moves beyond pairwise matching by learning view-consistent geometry and cross-view alignment cues that remain stable under wide baselines and limited overlap. When additional unposed assist views are available, \ours{} aggregates them via pose-graph canonical registration to increase geometric coverage and reinforce the final pose. Extensive experiments show that \ours{} achieves state-of-the-art performance across multiple benchmarks, substantially outperforming existing model-free methods, improving pose accuracy by \textbf{+12\%} on YCB-V and over \textbf{+20\%} on LM-O. Furthermore, \ours{} consistently performs well under both single-reference and sparse-reference settings, demonstrating strong robustness in real-world environments.

\end{abstract}

\section{Introduction}
With decades of progress in autonomous systems, modern embodied AI is increasingly expected to perceive, reconstruct, and orient previously unseen objects. A key requirement is estimating an object's 6D pose for downstream interaction such as grasping, placement, and tool use. However, many existing pipelines rely on strong prerequisites (e.g., accurate CAD models, object-specific assets, or dense posed image sequences for reconstruction), which significantly limits scalability and usability in open-world deployment.

We study 6D object pose estimation in a practical operational regime where supervision and assets are minimal at inference time. 
Given a query image and an \emph{arbitrary, unordered, and sparsely sampled} set of reference images of the same object instance, the goal is to recover the query pose in a canonical object-centric coordinate system, \emph{without CAD models and without explicit reconstruction}. 
From an information perspective, a single reference view often provides insufficient coverage of the object under occlusion and large viewpoint change (e.g., missing the back side), leading to intrinsic ambiguity in correspondence and pose.
Therefore, additional views can systematically reduce this uncertainty by increasing geometric coverage and establishing cross-view constraints. 
In our target setting, such extra observations come as \emph{pose-free assist views}: they are unposed, unconstrained snapshots whose camera poses are unknown and should not require annotation, yet they can act as intermediates to bridge wide-baseline gaps.

\begin{figure*}[t] 
    \hspace*{\fill} 
    \begin{minipage}[t]{\textwidth} 
        \centering
        \includegraphics[width=\textwidth]{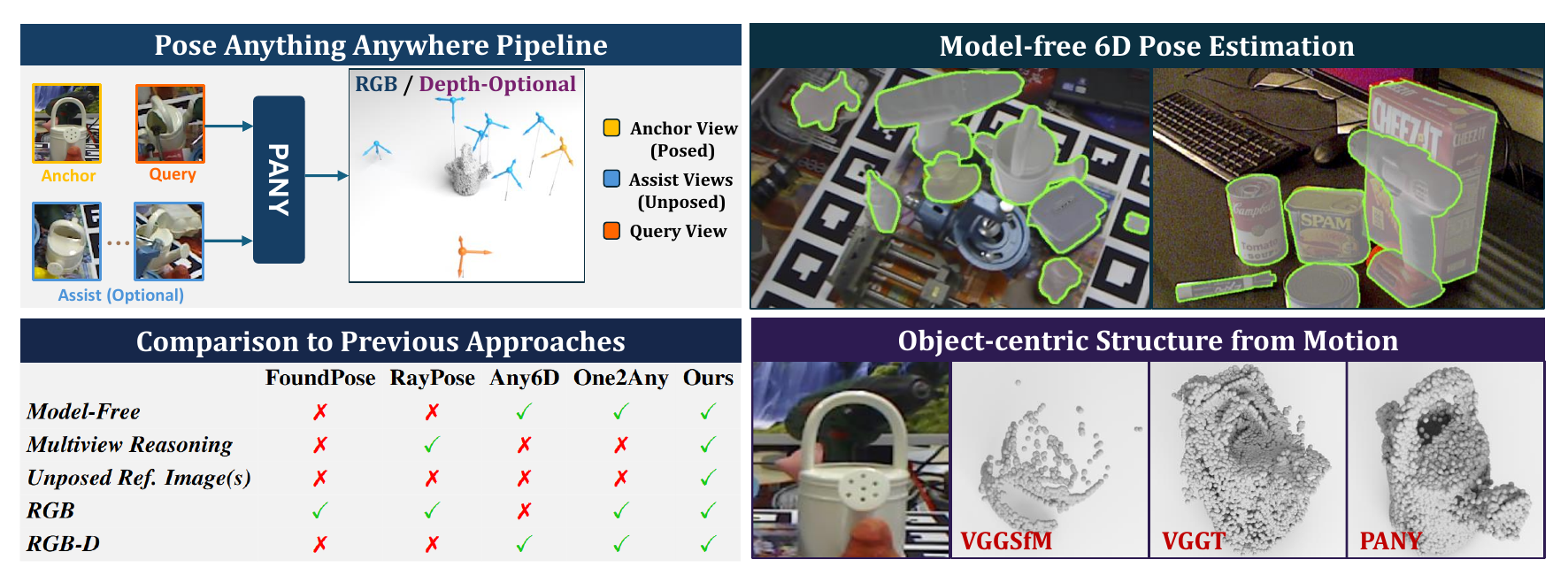} 
        \vspace{-0.5cm}
        \caption{ 
        \ours{} is the first model-free approach that simultaneously handles single- and sparse-view references and is compatible with RGB/RGBD input. 
        Compared to previous approaches~\cite{ornek2024foundpose,huang2025raypose,lee2025any6d,liu2025one2any}, \ours{} jointly infers object geometry and 6D pose from sparse and unposed reference views. It remains robust under large viewpoint changes and occlusions, and maintains stability in object-centric scenarios where VGGSfM~\cite{wang2024vggsfm} and VGGT~\cite{wang2025vggt} fail.
       }
        \vspace{-0.5cm}
        \label{fig:teaser}
    \end{minipage}
\end{figure*}

Existing approaches fall into three broad groups according to the reference modality: \textbf{(A) Model-based pipelines} rely on CAD models or object-specific 3D assets to generate templates or renderings for pose estimation~\cite{wang2021gdr,haugaard2022surfemb,ornek2024foundpose,nguyen2024gigapose,huang2024matchu,lin2024sam}. 
While highly accurate, they require curated 3D libraries and are less scalable for open-world objects.
\textbf{(B) Reconstruction-first pipelines} remove the need for CAD models by reconstructing an object from multiple views (e.g., SfM/NeRF/3DGS) and then applying a model-based solver~\cite{sun2022onepose,he2022onepose++,wen2024foundationpose,li2023nerf,lee2025any6d,matteo20246dgs,kerbl20233d}. Although effective, this direction typically incurs heavy capture or computation and does not directly support rapid onboarding from sparse, unordered references.
\textbf{(C) Pairwise model-free methods} estimate pose by matching a query image to a single reference via dense correspondences or learned pose inference~\cite{park2020latentfusion,wang2023posediffusion,lin2024relpose++,wang2026singref6d,liu2025one2any,corsetti2024open,geng2025one}. 
However, under large viewpoint change and occlusion, the query--reference overlap can be weak, making the problem intrinsically ambiguous and limiting robustness in the arbitrary sparse-reference regime. Recent geometry foundation models~\cite{wang2025vggt} provide strong multi-view geometry estimation from RGB inputs, but are trained with tracking-style supervision and often yield view-dependent representations, which are insufficient for stable \emph{object-centric canonical alignment} across wide baselines and unordered reference sets (Fig.~\ref{fig:teaser}).

Motivated by those observations, we present \textbf{\ours{}}, a model-free system for estimating object pose from arbitrary sparse references with optional pose-free assist views. In contrast to pairwise model-free approaches that rely on a single reference–query match, \ours{} performs multi-view reasoning across sparse and unordered observations to resolve pose ambiguity under occlusion and wide viewpoint changes.
Unlike reconstruction-first pipelines that require explicit object reconstruction before pose estimation, \ours{} directly infers object-centric canonical alignment from sparse views without reconstructing a full 3D model. \ours{} builds upon a geometry foundation model and adapts it from scene-level prediction to object-centric canonical alignment. 
To reduce the ambiguity induced by limited overlap, \ours{} learns geometry-aware cross-view correspondences that remain consistent under wide-baseline viewpoint changes and occlusion. 
When additional unposed views are available, \ours{} further aggregates them as intermediates via a pose-graph-based canonical registration, increasing geometric coverage and reinforcing the final pose even when no single reference overlaps well with the query.

Extensive experiments on several public benchmarks demonstrate the effectiveness of our framework design, enabling \ours{} to generalize across modalities (RGB/RGB-D) and reference configurations(single reference or arbitrary sparse references with pose-free assist views), setting a new state of the art on several challenging benchmarks. Our key contributions are:

\begin{itemize}
    \item We formalize pose estimation from \textbf{arbitrary sparse, unordered references} (optionally with pose-free assist views) as a practical operational regime, and propose \textbf{\ours{}} to address it without CAD models or onboarding stage.

    \item \ours{} adapts geometry foundation models to \textbf{object-centric canonical alignment} by learning geometry-aware, spatially consistent cross-view correspondences for robust pose reasoning under weak texture, occlusion, and wide baselines.

    \item \ours{} introduces an optional \textbf{multi-view inference} procedure that aggregates pose-free assist views via pose-graph canonical registration, improving robustness when no single reference view provides sufficient overlap.
\end{itemize}

\section{Related Work}
\label{sec:related}
Recently, model-based unseen object pose estimation methods have shown remarkable progress, achieving performance comparable to overfitting approaches on public benchmarks~\cite{hodan2024bop, sundermeyer2023bop}. With access to the CAD models, model-based approaches can generate a large set of templates to perform 2D-3D alignment~\cite{shugurov2022osop,ausserlechner2024zs6d,ornek2024foundpose,nguyen2024gigapose,moon2024genflow,moon2024genflow,huang2025raypose}, direct 3D-3D matching~\cite{chen2023zeropose,lin2024sam,huang2024matchu,caraffa2024freeze}, or adopt a render-and-compare strategy~\cite{labbe2022megapose,tremblay2023diff,li2018deepim,wen2024foundationpose} to refine the estimated pose iteratively. 
However, their reliance on accurate CAD models as references restricts their applicability in general daily scenarios where such models are unavailable. In this section, we review recent advances in \textbf{model-free novel object pose estimation}, focusing on approaches based on object reconstruction, relative pose estimation, and multiview reconstruction with foundational models.

\noindent \newline \textbf{Approaches based on object reconstruction.}
Instead of relying on accurate CAD models, some approaches reconstruct the 3D shape of the object from multi-view RGB images and align the input 2D image with the reconstructed 3D representation.
Reconstruction can be achieved through traditional structure-from-motion (SfM) pipelines~\cite{sun2022onepose,he2022onepose++,liu2022gen6d,he2022fs6d}, or implicitly using neural representations such as Neural Radiance Fields (NeRFs)~\cite{li2023nerf,wen2024foundationpose,wen2021bundletrack,wen2023bundlesdf,di2024zero123} and 3D Gaussian Splatting (3DGS)~\cite{matteo20246dgs,cai2025gs,jin20256dope}.
However, these methods typically require video sequences or dense sets of posed images and involve a lengthy onboarding process for each new object.
Recent studies aim to relax these constraints.
NOPE~\cite{nguyen2024nope} uses a single RGB image and matches it against synthetic templates generated from multiple viewpoints via a generative model.
GigaPose~\cite{nguyen2024gigapose} employs an image-to-3D model~\cite{long2024wonder3d} to recover object geometry, followed by model-based pose estimation protocols, but still requires size initialization. Any6D~\cite{lee2025any6d} further extends this idea by jointly estimating object scale and pose from only a single RGB-D reference image. 
OnePoseViaGen~\cite{geng2025one} uses a powerful image-to-3D framework to reconstruct high-quality object 3D models.
Although these methods take a single image as the anchor view, their pose estimation pipelines remain fundamentally model-based, relying on template matching against reconstructed object models. As a result, their performance is highly sensitive to the quality of the generated models, which can introduce geometric and appearance hallucinations when the object is only partially observed.

\noindent \newline \textbf{Approaches based on relative pose estimation.}
Extensive research~\cite{park2020latentfusion,wang2023posediffusion,sun2021loftr,zhang2022relpose,lin2024relpose++,zhang2024cameras,kim2025refpose} has addressed relative camera pose estimation between query and reference images using dense descriptor matching~\cite{sun2021loftr,sarlin2020superglue}, pose diffusion~\cite{zhang2024cameras,wang2023posediffusion}, or learning-based bundle adjustment~\cite{zhang2022relpose,lin2024relpose++}.
While effective for camera localization, these methods struggle to generalize to object-level pose estimation, where the target occupies smaller image regions and scale recovery is required.
To overcome this, model-free approaches directly predict relative pose from a reference RGB or RGB-D image without explicit 3D reconstruction.
Oryon~\cite{corsetti2024open} extends this paradigm to open-vocabulary settings, One2Any~\cite{lee2025any6d} predicts Reference Object Coordinate (ROC) maps from RGB-D pairs, and 3DAHV~\cite{zhao20243d}, DVMNet~\cite{zhao2024dvmnet}, and SingRef6D~\cite{wang2026singref6d} achieve single-image generalization through shared 2D–3D embeddings or pretrained matchers~\cite{sun2021loftr}.
Despite these advances, accurately reasoning geometric correlations under large viewpoint changes or discontinuities remains an open challenge.

\noindent \newline \textbf{Grounding models for visual geometry reasoning.}
Recent works have explored multi-view architectures to jointly learn geometry and semantics for reconstruction and pose estimation from unposed RGB inputs.
DUSt3R~\cite{wang2024dust3r} and MASt3R~\cite{leroy2024grounding} predict dense aligned point maps and camera parameters from uncalibrated image pairs.
Fast3R~\cite{yang2025fast3r} scales this idea to hundreds of unordered, unposed images through a transformer-based multi-view formulation, enabling efficient large-scale inference.
VGGT~\cite{wang2025vggt} further unifies multi-view camera pose, depth, point map, and correspondence prediction within a single framework.
Despite the impressive performance of grounded visual geometry models on scene-level inputs, scaling them to the object level remains challenging due to their view-dependent representations, especially when objects are only partially observed and exhibit significant viewpoint variation and occlusion, which hinder consistent geometric reasoning. Building on the strong geometry foundation model, our method eliminates view dependency in the predicted point maps and learns cross-view, spatially consistent 3D representations for robust pose alignment under large viewpoint changes and cluttered scenes.

\section{Method}
\begin{figure*}[t] 
    \hspace*{\fill} 
    \begin{minipage}[t]{\textwidth} 
        \centering
        \includegraphics[width=\textwidth]{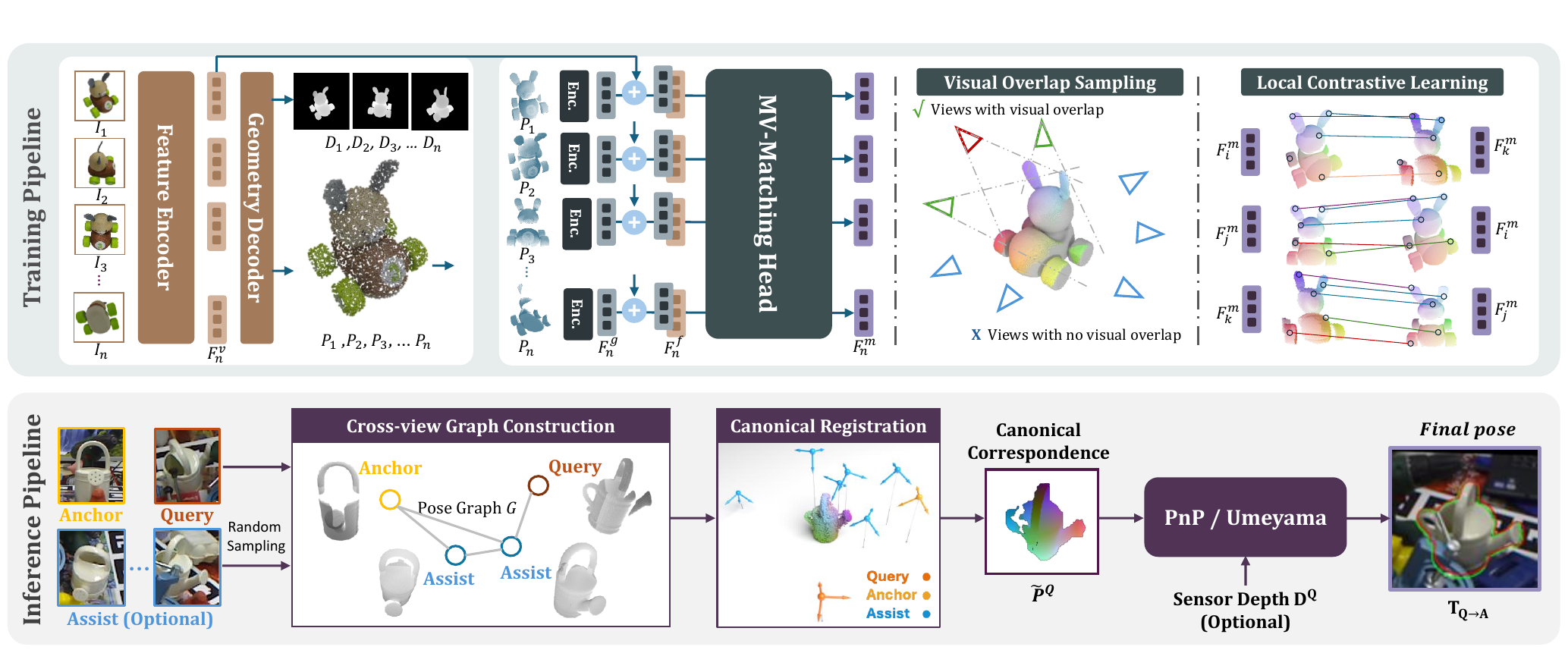} 
        \vspace{-0.5cm}
        \caption{ Overview of the PANY framework.
        (i) Given a query image and multiple reference views (at least one posed anchor image and optionally a few unposed assist views). Our network takes all views as input, jointly learns local geometry, dense 3D correspondences, and models spatial-consistent view representations for robust object pose estimation.
        (ii) During inference, the assist views act as intermediates to bridge limited visual overlap between the query and anchor views, enabling reliable 3D reasoning under large viewpoint variations.
       }
        \vspace{-0.5cm}
        \label{fig:method}
    \end{minipage}
\end{figure*}

Our goal is to estimate the 6D pose of an object from sparse reference observations without requiring CAD Models or onboarding stage. 
The key challenge arises when the query and reference views exhibit limited visual overlap due to occlusion or large viewpoint changes.
To address this challenge, we propose a canonical multi-view alignment framework that combines three key components: 
(1) a multi-view geometry backbone that predicts object-centric 3D structure from sparse views, 
(2) a cross-view correspondence learning module that establishes geometry-consistent matches across wide-baseline viewpoints, and 
(3) a pose-graph aggregation mechanism that integrates pose-free assist views to increase geometric coverage and resolve pose ambiguity.
Together, these components enable robust pose reasoning from arbitrary sparse references, even when no single reference sufficiently overlaps with the query.

\subsection{Pipeline Overview.}
Our method takes an arbitrary query view of an object and conditions on a sparse set of reference views, among which the anchor view defines the object's canonical pose, while the remaining assist views are pose-free.
As illustrated in Figure~\ref{fig:method}, the framework comprises two stages: a training pipeline and an inference pipeline.
During training, we employ a transformer backbone to predict dense point maps and jointly train a cross-view 3D matching head using contrastive learning on local point features.
By enforcing cross-view geometric constraints through dense 3D correspondences, the model learns spatially consistent and robust multi-view representations.
When the query and anchor views share little visual overlap, direct pose estimation becomes ambiguous. To resolve this ambiguity, we leverage pose-free assist views as intermediates that increase geometric coverage across viewpoints. We aggregate these views through
a pose-graph registration procedure that aligns local reconstructions into a consistent canonical coordinate frame at inference time.

\subsection{Task Formulation.}
As illustrated in Figure~\ref{fig:method}, given a RGB(-D) \textit{query image} denoted as $I^{\mathrm{Q}}$ and $K$ \textit{reference images} $\{ I^{\mathrm{ref}}_i \}_{i=1}^{K}$, our goal is to estimate the object pose $\mathbf{T}^Q$ in the query image. Among $\{ I^{\mathrm{ref}}_i \}_{i=1}^{K}$, one of which has a predefined canonical pose $\mathbf{T}^A$, termed the \textit{anchor view} $I^{\mathrm{A}}$, the remaining randomly captured pose-free reference views are denoted as \textit{assist views} $\{ I^{\mathrm{U}}_i \}_{i=1}^{M}$, which optionally providing complementary 3D contextual cues to enhance pose reasoning robustness. The query, anchor, and assist images form the complete set of inputs $\mathcal{I} = \{ I^{\mathrm{Q}} \} \cup \{ I^{\mathrm{A}} \} \cup \{ I^{\mathrm{U}}_i \}$, with the total number of views $N = 2 + M,M\ge0$. Our network takes $\mathcal{I}=\{I_i\}_{i=1}^N$ as input and estimates the relative object transformation between the query and anchor views $\mathbf{T}_{Q\rightarrow A}$ by leveraging the pose-free assist views. The final object pose in the query image can be solved by $\mathbf{T}^Q= \mathbf{T}^A \mathbf{T}^{-1}_{Q\rightarrow A}$.   

\subsection{Multi-view Geometric Reasoning}
Given $N$ input images $\{I_i\}_{i=1}^N$, each image is first patchified and encoded into tokens using a DINOv2~\cite{oquab2023dinov2} encoder. 
These tokens are then processed by alternating frame-wise intra-attention within each image and global cross-attention across all images, allowing the model to capture both local appearance features and cross-view geometric correlations. 
The resulting features for each image, termed as \textit{visual embedding} $\{ F^v_i \}_{i=1}^{N}$ are decoded using DPT-based upsampling heads to produce dense geometric predictions, including depth maps $\hat{D}_i \in \mathbb{R}^{H \times W}$ and point maps $\hat{P}_i \in \mathbb{R}^{3 \times H \times W}$.
\begin{equation}
    f: \{I_i\}_{i=1}^N \;\longrightarrow\; \{(\hat{D}_i, \hat{P}_i)\}_{i=1}^N.
\end{equation}

\subsection{Cross-view 3D Matching}
\label{sec:cross-view-matching}
Dense 3D point maps provide a bridge across views, yet the associated features remain view-dependent, making local matching unstable under wide-baseline changes. Without explicit cross-view consistency constraints, independently extracted embeddings may drift and lead to unreliable correspondences
To address these issues (Figure~\ref{fig:method}), we introduce a \textit{cross-view 3D matching model} with three components aligned with the following subsections:
(1) a geometry-aware feature encoding and fusion module that converts per-view point maps into 3D-aligned tokens and produces rotation-invariant descriptors;
(2) a local contrastive objective that supervises cross-view correspondences on the resulting 3D descriptors;
(3) an end-to-end joint training scheme that couples geometry prediction and correspondence learning to obtain consistent multi-view 3D representations.

\noindent \textbf{Cross-view 3D Matching Model.}
Given the predicted dense point map $\hat{P}_i$ for each view $I_i$, we encode local geometry using a geometric transformer backbone~\cite{lin2024sam,qin2022geometric}. 
A geometric encoder with rotation-invariant positional embedding extracts geometric features $F_i^{g}$ from $\hat{P}_i$. 
In parallel, per-pixel visual features $F_i^{v}$ are obtained from the visual backbone and lifted to 3D through coordinate indexing with $\hat{P}_i$. 
The two modalities are fused to form tokens $F_i^{f}$.

The fused tokens together with their 3D coordinates are processed by a geometry-aware matching head
\begin{equation}
F_i^{m} = \mathcal{T}_{\text{match}}\!\left(F_i^{f},\, \hat{P}_i\right),
\end{equation}
where $\mathcal{T}_{\text{match}}(\cdot)$ denotes a geometric transformer that alternates sparse geometric attention and dense linear attention~\cite{lin2024sam}. 
The resulting descriptors $F_i^{m}$ are rotation-invariant and geometry-aligned, enabling robust cross-view correspondence learning under viewpoint and appearance variations.

\noindent \textbf{Local Contrastive Learning.}
Wide-baseline viewpoint changes often produce view-dependent representations that hinder reliable cross-view correspondence. 
To address this issue, we train the matching head using local contrastive supervision defined on the geometry-aware descriptors.
Given a set of $K$ input views with point sets $\{P_i\}_{i=1}^{K}$ and descriptors $\{F_i^{m}\}_{i=1}^{K}$, we compute cross-view attention matrices between view pairs as
\begin{equation}
A_{ij} = F_i^{m} (F_j^{m})^{\top},
\end{equation}
where $A_{ij} \in \mathbb{R}^{N_i \times N_j}$ measures descriptor similarity between points from views $i$ and $j$.
For each valid view pair $(i,j)$, the attention matrix is supervised using a bidirectional cross-entropy objective
\begin{equation}
\mathcal{L}_{ij}
=
\mathrm{CE}(A_{ij}, \hat{Y}_i)
+
\mathrm{CE}(A_{ij}^{\top}, \hat{Y}_j),
\end{equation}
where $\hat{Y}_i$ and $\hat{Y}_j$ denote the ground-truth correspondences. 
The final matching loss aggregates all valid pairs
\begin{equation}
\mathcal{L}_{\text{match}} =
\frac{1}{|\mathcal{E}|}
\sum_{(i,j)\in\mathcal{E}}
\mathcal{L}_{ij}.
\end{equation}
Since the predicted point maps lie in a shared coordinate frame, the correspondence label for a point $p_i \in P_i$ is defined by nearest-neighbor search
\begin{equation}
k^* = \arg\min_k \|p_i - p_{j,k}\|_2 ,
\end{equation}
and
\begin{equation}
y_i =
\begin{cases}
0, & \|p_i - p_{j,k^*}\|_2 \ge \delta_{\text{dis}} \\
k^*, & \text{otherwise},
\end{cases}
\end{equation}
where $\delta_{\text{dis}}$ is a distance threshold that determines whether a valid correspondence exists.

\noindent \textbf{Joint Learning for Geometry and Matching}
To jointly enable accurate 3D geometry prediction and cross-view correspondence learning from multi-view RGB inputs, We jointly optimize the geometry prediction module and the correspondence learning module
in an end-to-end manner.
Our training objective combines supervision for both geometric reconstruction and cross-view contrastive learning. 
The geometry head predicts dense depth maps $\hat{D}_i$ and point maps $\hat{P}_i$, which are supervised using $\ell_1$ regression losses against their ground-truth counterparts:
\begin{equation}
    \mathcal{L}_{\text{depth}} = \frac{1}{N} \sum_{i=1}^{N} \| \hat{D}_i - \overline{D}_i \|_1,
    \quad
    \mathcal{L}_{\text{point}} = \frac{1}{N} \sum_{i=1}^{N} \| \hat{P}_i - \overline{P}_i \|_1.
\end{equation}
For cross-view correspondence supervision, we adopt the local contrastive loss described in Section~\ref{sec:cross-view-matching}. 
The final objective integrates all components:
\begin{equation}
    \mathcal{L} = 
    \lambda_d \, \mathcal{L}_{\text{depth}} +
    \lambda_p \, \mathcal{L}_{\text{point}} +
    \lambda_m \, \mathcal{L}_{\text{match}},
\end{equation}
where $\lambda_d$, $\lambda_p$, and $\lambda_m$ are balancing weights. 
This joint objective enforces accurate per-view geometry prediction while learning geometry-aware embeddings that remain consistent across multiple views.

\subsection{Addressing Large-view Gap}
Large viewpoint gaps between the query and anchor views often lead to limited visual overlap, making direct pose estimation unreliable. 
To mitigate this issue, we leverage pose-free assist views as intermediate observations that increase geometric coverage across viewpoints. 
Our pipeline is trained with multi-view inputs, enabling flexible inference with an arbitrary number of views and ensuring 3D-consistent point map predictions across them.

\noindent \textbf{Pose Graph Construction.}
Given a query view $I^Q$, an anchor view $I^A$ with canonical pose $\mathbf{T}^A$, and $M$ pose-free assist views $\{I_i^U\}_{i=1}^M$, the network predicts dense point maps $\hat{P}_i$ and cross-view correspondences across all views. 
We construct a pose graph $\mathcal{G}=(\mathcal{V},\mathcal{E})$, where each node represents a view and its predicted point cloud. 
An edge $(i,j)\in\mathcal{E}$ is created if the number of valid correspondences between two views exceeds a threshold $\eta$. 
For each connected pair, we estimate a similarity transformation $\mathbf{T}_{i\leftarrow j}$ using robust point-set alignment with RANSAC and Umeyama estimation.

\noindent \textbf{Canonical Alignment.}
To recover a globally consistent reconstruction, we align all predicted point clouds into the anchor-centric coordinate frame. 
Starting from the anchor node, we traverse the pose graph and compose the estimated transformations along the graph path:
\begin{equation}
\mathbf{T}_{i\rightarrow A} =
\prod_{(k,l)\in\pi(i,A)} \mathbf{T}_{k\leftarrow l}.
\end{equation}
The predicted point clouds are then registered into the canonical frame as
\[
\tilde{P}_i=\mathbf{T}_{i\rightarrow A}\hat{P}_i .
\]
This multi-view alignment aggregates geometric information from multiple views and resolves pose ambiguity that cannot be addressed by pairwise matching alone.

\noindent \textbf{Final Pose Estimation.}
Once the query point cloud is registered into the canonical coordinate system, the final object pose in the query image is recovered as
\begin{equation}
\mathbf{T}^Q = \mathbf{T}^A \mathbf{T}_{Q\rightarrow A}^{-1}.
\end{equation}
The relative transformation $\mathbf{T}_{Q\rightarrow A}$ is estimated from the canonical point clouds using Kabsch--Umeyama alignment when depth is available, or via a PnP optimization using predicted 3D--2D correspondences for RGB-only inputs.

\section{Experiments}
\subsection{Implementation Details}
Input images are pre-processed by cropping and resizing to a resolution of 518$\times$518. During training, the VGGT~\cite{wang2025vggt} feature backbone is frozen; we apply LoRA ~\cite{hu2022lora} with a rank of 8 and an alpha value of 32 to the frozen components.  The proposed matching head is trained from scratch, jointly with the LoRA finetuning. We set the loss weights to $\lambda_d= 1.0$, $\lambda_p = 1.0$, and $\lambda_m = 0.05$ for balancing geometry and matching objectives. The contrastive InfoNCE temperature is set to $\tau = 0.1$. For each training sample, given a query image, we randomly select 2 to 8 reference views, resulting in a total of 3 to 9 input images per sample. Among these references, at least two are required to be positive reference views, with the positive view threshold $\theta_{\text{OFP}}=10^\circ$. 
For cross-view graph construction, we set the minimum correspondence 
threshold to $\eta = 500$ valid 3D pairs from 2048 samples per view pair.
The model is trained for 15 epochs with a batch size of 16, where each epoch consists of 500 iterations. We train the model on two NVIDIA A100 GPUs for approximately 36 hours. We use the AdamW optimizer ~\cite{kingma2014adam} with a learning rate of 1e-4 and employ a warm-up strategy during the first epoch. 

\subsection{Dataset}
\noindent \textbf{Training dataset.} 
To enhance the generalization ability of our model to diverse real-world objects, we train our model on the large-scale synthetic Omni6DPose dataset~\cite{zhang2024omni6dpose}, 
which contains more than $5{,}000$ CAD objects spanning 149 categories. For training, we extract object-centric crops and construct paired reference-query samples, resulting in over 2 million RGB-D images with accurate ground-truth poses. 

\noindent \textbf{Test benchmarks.} To evaluate the robustness and generalization ability of our method, we test our method on 5 real-world datasets that have never been seen during training. 
We report results on LINEMOD~\cite{hinterstoisser2011multimodal}, 
YCB-Video~\cite{xiang2017posecnn},
Toyota-Light~\cite{hodan2018bop}, 
LM-O~\cite{brachmann2014learning}, 
and Real275~\cite{wang2019normalized} test set. 

\noindent \textbf{Evaluation metrics.}
We follow the BOP benchmark~\cite{hodan2018bop} and previous methods~\cite {corsetti2024open,lee2025any6d,liu2025one2any,ornek2024foundpose} to evaluate the pose estimation performance on single-reference and sparse-reference setups. For fair comparison with the previous methods, we adopt the Average Recall (\textbf{AR}), \textbf{ADD-0.1d}, and \textbf{ADD AUC} as our evaluation metrics.  

\subsection{Experiment Setup}
To demonstrate the flexibility and scalability of our approach, we compare against recent state-of-the-art methods under varying input configurations. Following Oryon~\cite{corsetti2024open}, which samples image pairs for testing, we evaluate our method alongside RGB~\cite{wang2023posediffusion,lin2024relpose++,park2020latentfusion,liu2025one2any,wang2026singref6d} and RGB-D~\cite{gumeli2023objectmatch,lin2022category,lee2025any6d,liu2025one2any} baselines on the Real275~\cite{wang2019normalized} and Toyota-Light~\cite{hodan2018bop} datasets. We also follow the One2Any~\cite{liu2025one2any} protocol that directly uses the first view as the anchor image, evaluating our method on LINEMOD~\cite{hinterstoisser2011multimodal} and YCB-Video~\cite{xiang2017posecnn}. For the multi-reference setup, we further compare with recent template-based methods on the LM-O~\cite{brachmann2018learning} dataset.

\subsection{Comparison to Single-Reference Methods}

\begin{table}[t]
\centering
\begin{minipage}{0.75\linewidth}
\centering
\caption{Comparison with state-of-the-art methods on \textbf{Real275}~\cite{wang2019normalized} and \textbf{Toyota-Light}~\cite{hodan2018bop} datasets. \textbf{All the methods use the same query-reference image pairs}.} \vspace{-0.2cm}
\label{tab:real_toyota}
\resizebox{\columnwidth}{!}{%
\begin{tabular}{l|c|cc|cc}
\toprule
\multirow{2}{*}{\textbf{Method}} & \multirow{2}{*}{\textbf{Mod.}} &
\multicolumn{2}{c|}{\textbf{Real275}~\cite{wang2019normalized}} &
\multicolumn{2}{c}{\textbf{Toyota-Light}~\cite{hodan2018bop}} \\
 & & AR & ADD-0.1d & AR & ADD-0.1d \\
\midrule
PoseDiffusion~\cite{wang2023posediffusion} & RGB & 9.2 & 0.8 & 7.8 & 1.2 \\
RelPose++~\cite{lin2024relpose++}          & RGB & 22.8 & 11.9 & 30.9 & 11.6 \\
LatentFusion~\cite{park2020latentfusion}   & RGB & 22.6 & 9.6 & 28.2 & 10.2 \\
One2Any~\cite{park2020latentfusion}        & RGB & 26.6 & 4.8 & 22.7 & 3.7 \\
SingRef6D~\cite{wang2026singref6d} & RGB & 28.7 & 11.6 & 31.7 & \textbf{13.1} \\

\rowcolor{gray!20} \textbf{\ours{} (Ours)}  & RGB & \textbf{59.3} & \textbf{27.8} & \textbf{37.5} & 10.8 \\
\midrule
ObjectMatch~\cite{gumeli2023objectmatch}   & RGBD & 26.0 & 13.4 & 9.8  & 5.4 \\
Oryon~\cite{corsetti2024open}              & RGBD & 46.5 & 34.9 & 34.1 & 22.9 \\
Any6D~\cite{lee2025any6d}                  & RGBD & 51.0 & 53.5 & 43.3 & 32.2 \\
One2Any~\cite{liu2025one2any}              & RGBD & 54.9 & 41.0 & 42.0 & 34.6 \\
\rowcolor{gray!20} \textbf{\ours{} (Ours)}  & RGBD & \textbf{81.8} & \textbf{89.3} & \textbf{51.1} & \textbf{39.8} \\
\bottomrule
\end{tabular}%
}
\end{minipage}

\vspace{2em}

\begin{minipage}{0.8\linewidth}
\centering
\caption{
Pose estimation performance on \textbf{YCB-Video}~\cite{xiang2017posecnn} 
and \textbf{LINEMOD}~\cite{esposito2016multimodal} datasets.
\textbf{All approaches use the first image from each scene as the reference view.} Methods marked with * require additional CAD models or ground-truth translations.}
\label{tab:ycbv_lm}
\resizebox{\columnwidth}{!}{%
\setlength{\tabcolsep}{6pt}
\begin{tabular}{l|c|c|c}
\toprule
\textbf{Method} & \textbf{Ref. Images} & 
\makecell{\textbf{YCB-V}~\cite{xiang2017posecnn}\\ADD AUC} & 
\makecell{\textbf{LINEMOD}~\cite{esposito2016multimodal}\\ADD-0.1d} \\
\midrule
FS6D~\cite{he2022fs6d} + ICP & 16 & 42.1 & 91.5 \\
\midrule
FoundationPose*~\cite{wen2024foundationpose} & 1 (CAD) & 76.1 & 48.3 \\
NOPE~\cite{nguyen2024nope}* & 1 + GT Trans & 25.1 & 16.3 \\
Oryon~\cite{corsetti2024open} & 1 & 7.4 & 9.8 \\
One2Any~\cite{liu2025one2any} & 1 & 84.4 & 52.6 \\
\rowcolor{gray!20} 
\textbf{\ours{} (Ours)} & 1 & 92.5 & 55.3 \\ 
\midrule
\rowcolor{gray!20}
\textbf{\ours{} (Ours)} & 1 + 8 unposed & \textbf{94.6} & \textbf{87.4} \\
\bottomrule
\end{tabular}}
\end{minipage}
\end{table}

\textbf{Evaluation on Real275 and Toyota-Light datasets.} We compare our approach with baselines that only take a single RGB or RGB-D reference image as input on the Real275~\cite{wang2019normalized} and Toyota-Light~\cite{hodan2018bop} datasets, as summarized in Table~\ref{tab:real_toyota}. 
For fair comparison, we follow the evaluation protocol established by Oryon~\cite{corsetti2024open}, and evaluate 2{,}000 query–reference image pairs per test set, assuming ground-truth masks are available. 
As shown in Table~\ref{tab:real_toyota}, our method consistently outperforms all alternative approaches across all metrics when only a single reference view is provided. 
Specifically, compared to methods that estimate depth from the input RGB image in SingRef6D~\cite{wang2026singref6d}, generate 3D geometry from images in Any6D~\cite{lee2025any6d}, or perform point cloud registration from visual embeddings in Oryon~\cite{corsetti2024open}, our model achieves more accurate query-reference alignment even with a single reference image, demonstrating the robustness and generalizability of our 3D reasoning pipeline.

\begin{figure}[t]
    \centering
    \includegraphics[width=1.0\textwidth]{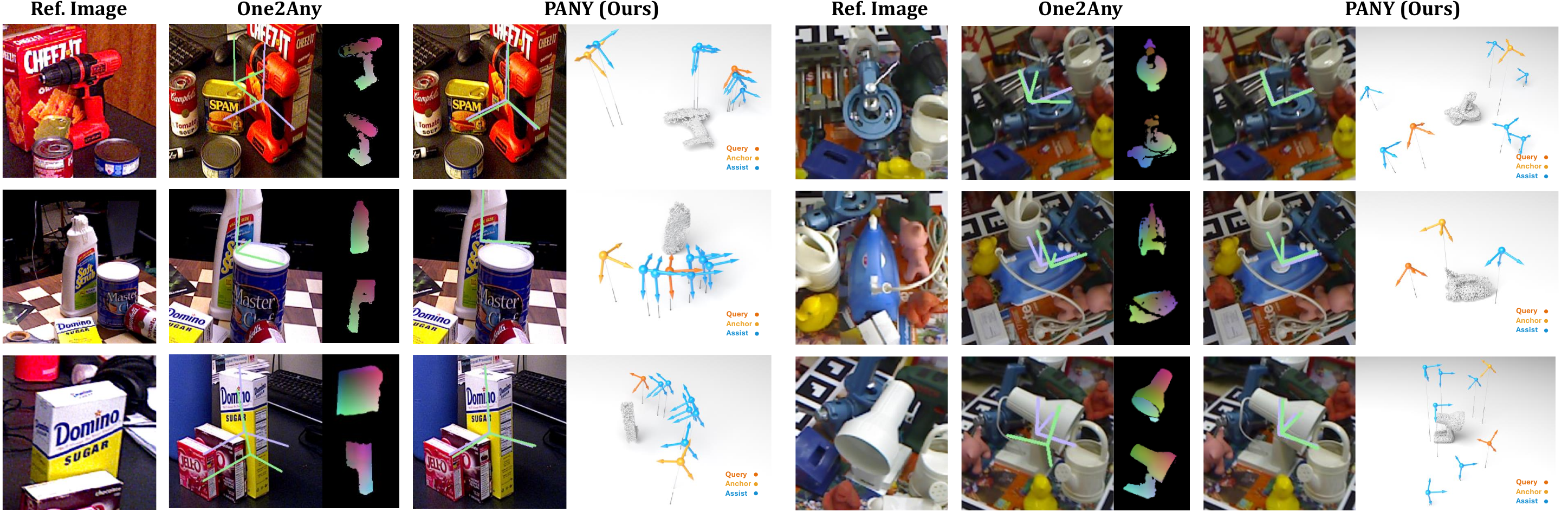}\vspace{-0.3cm}
    \caption{Quantative Comparison between One2Any~\cite{liu2025one2any} and Ours. 
    The predicted poses are displayed in green and ground truth poses are in pink. The second column shows the pose estimation results of One2Any~\cite{liu2025one2any} through pair-wise prediction; the last column shows how we leverage unposed references to conduct multi-view reasoning under large viewpoint changes.}
    \vspace{-0.4cm}
    \label{fig:vis_ycblm}
\end{figure}

\noindent \textbf{Evaluation on YCB-V and LINEMOD datasets.} 
Following One2Any~\cite{liu2025one2any}, we evaluate our method on the occluded YCB-V~\cite{xiang2017posecnn} and LINEMOD~\cite{hinterstoisser2011multimodal} datasets, using only the first view as the posed reference and assuming available ground-truth masks. YCB-V poses challenges due to severe occlusions, while LINEMOD contains textureless and irregular objects captured in nearly 360° video sequences, making single-view reference particularly difficult. We adopt FoundationPose~\cite{wen2024foundationpose} results from~\cite{liu2025one2any} for comparison. Our \ours{} (single-ref.) uses only one anchor image as reference, whereas \ours{} (multi-ref.) incorporates eight unposed assist views. Table~\ref{tab:ycbv_lm} shows that our method outperforms all baselines, and qualitative results in Figure~\ref{fig:vis_ycblm} highlight robustness to occlusion, textureless surfaces, and large viewpoint changes. Those experiments demonstrate that incorporating assist views significantly boosts performance, validating the effectiveness of our multi-view reasoning pipeline under challenging conditions.

\subsection{Performance of Multi-view Pose Reasoning}
\begin{table}[!b]
\centering
\caption{Evaluation on \textbf{LM-O}~\cite{hinterstoisser2011multimodal} dataset compared to model-based methods with RGB image inputs and all methods use CNOS~\cite{nguyen2023cnos} mask as segmentation mask.}\vspace{-0.2cm}
\centering
\label{tab:coarse_lmo}
\resizebox{0.75\columnwidth}{!}{%
\begin{tabular}{l|c|c|c|c}
\toprule
\textbf{Method} & \textbf{\# Posed Ref.} & \textbf{Model-free} & \textbf{~~AR~~}&\textbf{Time(s)} \\
\midrule
MegaPose~\cite{labbe2022megapose} & 520 & \xmark & 22.9 & 2.53 \\
GenFlow~\cite{moon2024genflow} & 208 & \xmark & 25.0 & - \\
GigaPose~\cite{nguyen2024gigapose} & 162 & \xmark & 29.9 & 11.53 \\
FoundPose ~\cite{ornek2024foundpose} & 57 & \xmark & 39.7 & 1.7\\
RayPose ~\cite{huang2025raypose}& 8 & \xmark & 42.1 & \textbf{0.71}\\
\midrule
FoundPose~\cite{ornek2024foundpose} & 8 & \cmark & 34.7 & 1.7  \\
\rowcolor{gray!20} \textbf{\ours{} (Ours)} & 1 + 8 unposed & \cmark & \textbf{42.3} & \textbf{0.75} \\
\midrule
One2Any~\cite{liu2025one2any} & 1 & \cmark & 18.2 & \textbf{0.09}\\
Any6D~\cite{lee2025any6d} & 1 & \cmark & 28.6 & -\\
\rowcolor{gray!20} \textbf{\ours{} (Ours)} & 1 & \cmark & 
\textbf{36.9} &0.19\\
\bottomrule
\end{tabular}}
\vspace{-0.2cm}
\end{table}

\begin{figure}[!t]
    \centering
    \includegraphics[width=1.0\textwidth]{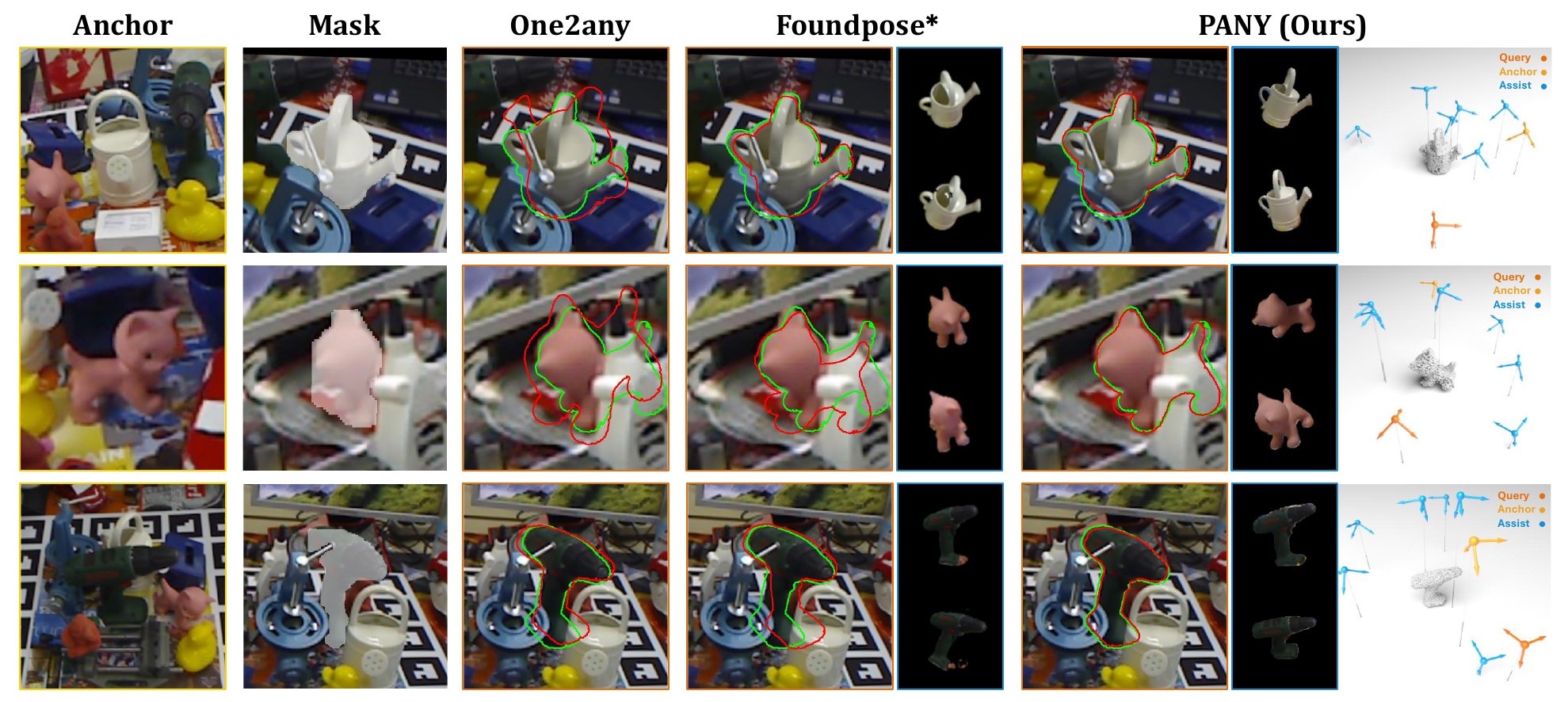}\vspace{-0.3cm}
    \caption{
    Qualitative comparison between One2Any~\cite{liu2025one2any}, FoundPose~\cite{ornek2024foundpose}, and \ours{}. 
    \textcolor{green}{Green} and \textcolor{red}{red} contours indicate ground-truth and predicted poses. 
    \textcolor{yellow!60!brown}{Yellow}, \textcolor{orange}{orange}, and \textcolor{cyan!50}{blue} mark the anchor view, query view, and assist views. FoundPose* additionally uses assist views with pose annotations.
    }
    
    \vspace{-0.3cm}
    \label{fig:lmo}
\end{figure}

\label{sec:Performance of Multi-view Pose Reasoning}
To evaluate the effectiveness of our multi-view pose reasoning, we conduct experiments on the challenging LM-O dataset~\cite{brachmann2014learning}, which features severe occlusions and large viewpoint changes. 
Following prior model-based unseen object pose estimation works, we use CNOS~\cite{nguyen2023cnos} masks instead of ground-truth segmentation. 
For each LM-O object, we select eight sparse reference views from LINEMOD~\cite{hinterstoisser2011multimodal}, with only the first view defining the canonical pose. 
We compare against state-of-the-art model-based RGB-only methods in Figure~\ref{fig:lmo}, and also reproduce FoundPose~\cite{ornek2024foundpose} under the same setting with additional in-plane rotations for fairness. 
As shown in Table~\ref{tab:coarse_lmo}, our method surpasses all model-based baselines that rely on dense CAD templates, despite using only eight posed anchors. 
Compared to One2Any~\cite{liu2025one2any}, which uses a single reference, our approach achieves nearly double the accuracy, demonstrating the strength of our multi-view reasoning and geometry-aligned design.

\subsection{Ablation Study}
\noindent \textbf{Model design.}
We analyze the contribution of each component of our canonical multi-view alignment framework on the Real-275~\cite{wang2019normalized} dataset under a controlled setting with one anchor and one query image (no assist views). Results are summarized in Table~\ref{tab:model_design}. 
\textit{(1) Geometry-only baseline.} 
As a minimal baseline, we directly align the predicted query and anchor point maps without explicit correspondence learning. This setting evaluates whether geometry prediction alone is sufficient for reliable pose recovery.
\textit{(2) Visual correspondence without geometric fusion.} 
We then introduce a cross-view matching module operating on visual embeddings only, without incorporating explicit geometric features. This improves setup (1) by 7.6\% in ADD-0.1d, indicating that learning cross-view correspondences already provides stronger alignment cues than geometry prediction alone.
\textit{(3) Full model.} 
Our complete framework further fuses geometric point features with visual embeddings to jointly learn geometry-aware correspondences in a shared 3D feature space. This design yields a 17.8\% improvement over the baseline.

\begin{table*}[!htbp]
\centering
\caption{\textbf{Controlled ablation under identical inference settings.} 
(a) Backbone-only baselines: replacing the geometry backbone (DUSt3R/MASt3R/VGGT) while keeping the same solver and without matching or aggregation is insufficient to match our performance. 
(b) Model design ablation on Real275: under identical single-anchor settings (no assist views), progressively adding matching and geometric fusion yields consistent gains, demonstrating that improvements are not explained by backbone replacement alone.}
\vspace{-2mm}
\begin{subtable}[t]{0.38\textwidth}
\centering
\resizebox{\linewidth}{!}{
\begin{tabular}{lcc}
\toprule
Method & AR & ADD-0.1d \\
\midrule
vanilla DUSt3R & 27.8 & 16.8 \\
vanilla MASt3R & 29.2 & 17.6 \\
vanilla VGGT   & 43.6 & 33.4 \\
\midrule
Ours Full       & \textbf{51.1} & \textbf{39.8} \\
\bottomrule
\end{tabular}}
\caption{Backbone-only baselines.}
\label{tab:backbone_compare}
\end{subtable}
\hfill
\begin{subtable}[t]{0.58\textwidth}
\centering
\resizebox{\linewidth}{!}{
\begin{tabular}{lcc}
\toprule
Design Variant & AR & ADD-0.1d \\
\midrule
Geometry-only & 56.6 (\textcolor{green!50!black}{-4.6\%}) & 23.6 (\textcolor{green!50!black}{-15.1\%}) \\
+ Matching & 57.6(\textcolor{green!50!black}{-2.9\%}) & 25.4 (\textcolor{green!50!black}{-8.6\%})  \\
Full (Matching+Fusion) & \textbf{59.3} & \textbf{27.8} \\
\bottomrule
\end{tabular}}
\caption{Model design ablation.}
\label{tab:model_design}
\end{subtable}
\vspace{-4mm}
\end{table*}

Overall, these results show that the gains primarily come from the proposed cross-view correspondence learning and geometry-aware feature fusion, which together improve the spatial consistency of predicted geometry and enable more reliable pose alignment.

\noindent \textbf{Backbone-only baseline.}
To verify that the gains do not simply come from using a stronger geometry backbone, we evaluate backbone-only variants using DUSt3R, MASt3R, and VGGT to predict geometry while keeping the same downstream solver. 
As shown in Table~\ref{tab:backbone_compare}, replacing the backbone alone is insufficient, and our full system consistently outperforms all backbone-only baselines.

\pgfplotsset{compat=1.18}

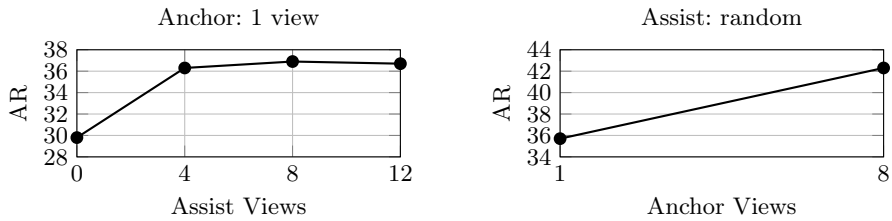
\begin{figure}[t]
\centering

\begin{minipage}{0.48\columnwidth}
\centering
\begin{tikzpicture}
\begin{axis}[
width=\linewidth,
height=3cm,
xlabel={Assist Views},
ylabel={AR},
xmin=0, xmax=12,
xtick={0,4,8,12},
ymin=28, ymax=38,
grid=both,
title={Anchor: 1 view}
]
\addplot[mark=*, thick] coordinates {
(0,29.8)
(4,36.3)
(8,36.9)
(12,36.7)
};
\end{axis}
\end{tikzpicture}
\end{minipage}
\hfill
\begin{minipage}{0.48\columnwidth}
\centering
\begin{tikzpicture}
\begin{axis}[
width=\linewidth,
height=3cm,
xlabel={Anchor Views},
ylabel={AR},
xmin=1, xmax=8,
xtick={1,8},
ymin=34, ymax=44,
grid=both,
title={Assist: random}
]
\addplot[mark=*, thick] coordinates {
(1,35.7)
(8,42.3)
};
\end{axis}
\end{tikzpicture}
\end{minipage}

\vspace{-0.2cm}
\caption{Impact of assist views and anchor views on LM-O.}
\label{fig:lmo_ablation}
\vspace{-0.4cm}

\end{figure}

\noindent \textbf{Effect of anchor view sampling.}
We evaluate robustness to anchor selection on LM-O~\cite{brachmann2014learning} by randomly sampling anchor images three times per object. The resulting AR scores remain nearly identical, indicating that the method is insensitive to anchor choice.

\noindent \textbf{Effect of the number of assist views.}
We vary the number of assist views during inference on LM-O and average over three random samples per setting. Performance improves with additional views and saturates beyond four, motivating our default use of eight assist views.

\noindent \textbf{Limitations.}
Our method may struggle in cases with extreme occlusion or very limited query reference overlap, where reliable correspondences cannot be established. Strongly symmetric objects can also introduce pose ambiguity when geometric cues alone are insufficient to resolve orientation.

\section{Conclusions} 
\label{sec:conclusion}
We present \ours{}, a scalable framework for model-free pose estimation from sparse RGB references. 
By jointly learning geometry-aware representations and cross-view correspondences, \ours{} enables robust object-centric alignment under large viewpoint changes and limited visual overlap. 
Experiments across multiple benchmarks demonstrate strong generalization across objects and reference configurations. 
Overall, \ours{} provides a practical step toward robust multi-view reasoning for real-world robotic perception.




%
%
\bibliographystyle{splncs04}
\bibliography{main}
\end{document}